\documentclass[final,5p,times,twocolumn]{elsarticle}

\usepackage{amssymb}
\usepackage{amsmath}
\usepackage{url}
\usepackage{algorithmic}
\usepackage{algorithm}
\usepackage{array}
\usepackage[caption=false,font=normalsize,labelfont=sf,textfont=sf]{subfig}
\usepackage{textcomp}
\usepackage{stfloats}
\usepackage{url}
\usepackage{graphicx}
\usepackage{verbatim}
\usepackage{bm}
\usepackage{cases}
\usepackage{multirow}
\usepackage{multicol}
\usepackage{booktabs}
\usepackage{chngcntr}
\usepackage{pifont}
\usepackage{color}
\usepackage[table]{xcolor}

\usepackage{adjustbox}

\journal{ISPRS Journal of Photogrammetry and Remote Sensing}

\begin{document}

\begin{frontmatter}



\title{Joint-Optimized Unsupervised Adversarial Domain Adaptation in Remote Sensing Segmentation with Prompted Foundation Model}


\author[1]{Shuchang Lyu}
\author[1]{Qi Zhao\corref{*}}
\ead{zhaoqi@buaa.edu.cn}
\author[3]{Guangliang Cheng}
\author[3]{Yiwei He}
\author[1]{Zheng Zhou}
\author[1]{Guangbiao Wang\corref{*}}
\ead{wanggb@buaa.edu.cn}
\author[2]{Zhenwei Shi}

\cortext[*]{Corresponding author}

\affiliation[1]{organization={School of Electronics and Information Engineering, Beihang University},
            city={Beijing},
            postcode={100191}, 
            country={China}}

\affiliation[2]{organization={ Image Processing Center, School of Astronautics, Beihang University},
            city={Beijing},
            postcode={100191}, 
            country={China}}
            
\affiliation[3]{organization={Department of Computer Science, University of Liverpool},
            city={Liverpool},
            postcode={L69 3BX}, 
            country={UK}}

\begin{abstract}
Unsupervised Domain Adaptation for Remote Sensing Semantic Segmentation (UDA-RSSeg) addresses the challenge of adapting a model trained on source domain data to target domain samples, thereby minimizing the need for annotated data across diverse remote sensing scenes. This task presents two principal challenges: (1) severe inconsistencies in feature representation across different remote sensing domains, and (2) a domain gap that emerges due to the representation bias of source domain patterns when translating features to predictive logits. To tackle these issues, we propose a \textbf{j}oint-\textbf{o}ptimized \textbf{a}dversarial \textbf{net}work incorporating the ``Segment Anything Model (SAM)'' (SAM-JOANet) for UDA-RSSeg. Our approach integrates SAM to leverage its robust generalized representation capabilities, thereby alleviating feature inconsistencies. We introduce a finetuning decoder designed to convert SAM-Encoder features into predictive logits. Additionally, a feature-level adversarial-based prompted segmentor is employed to generate class-agnostic maps, which guide the finetuning decoder’s feature representations. The network is optimized end-to-end, combining the prompted segmentor and the finetuning decoder. Extensive evaluations on benchmark datasets, including ISPRS (Potsdam/Vaihingen) and CITY-OSM (Paris/Chicago), demonstrate the effectiveness of our method. The results, supported by visualization and analysis, confirm the method's interpretability and robustness. The code of this paper is available at \url{https://github.com/CV-ShuchangLyu/SAM-JOANet/}.
\end{abstract}



\begin{keyword}
Unsupervised domain adaptation \sep Adversarial learning \sep Prompted foundation model \sep Semantic segmentation \sep Remote sensing
\end{keyword}

\end{frontmatter}



\section{Introduction}
\label{sec1}
Remote sensing semantic segmentation (RSSeg) has found extensive application across a range of real-world scenarios. A variety of advanced methods ~\cite{BSNet, Maoetal-Urban, SAMRS, aaai_RSSeg} have been developed to enhance its performance. However, despite these advances, the efficacy of RSSeg remains highly dependent on the similarity between training (source) and testing (target) datasets. Significant discrepancies between these datasets can markedly degrade performance. To address this challenge and facilitate knowledge transfer across domains, the unsupervised domain adaptation remote sensing semantic segmentation (UDA-RSSeg) task has emerged as a critical area of research.
\par In the UDA-RSSeg task, domain shift in RS images primarily arises from differences in ground sampling distance, variations in remote sensing sensors, and diverse geographical landscapes~\cite{ST-DASegNet}. To address these challenges, several approaches ~\cite{DA-DualGAN, Maetal-MBATA, Wangetal-FGUDA, DA-Road} employ adversarial learning to align source and target features. Additionally, other methods ~\cite{ST-UDA1, ST-UDA2, Crots} leverage self-training mechanisms to generate high-quality pseudo-labels for target annotations. Despite significant advancements, two critical issues persist: (1) Remote sensing images from different domains exhibit severe inconsistencies in feature representation. While adversarial learning aids in feature alignment, it does not fundamentally enhance generalized feature representation. (2) Even with improved feature representation, the tendency of source patterns to dominate can introduce domain gaps when mapping features to predictive logits.
\par To address these two issues, we propose SAM-JOANet, a joint-optimized adversarial network built on the prompted foundation model (SAM)~\cite{SAM}. For the first problem, we integrate SAM into our architecture to leverage its generalized representation capabilities and mitigate feature inconsistency. Given its extensive training on ``SA-1B'', SAM excels at representing images from various domains. We also design a fine-tuning decoder to map features from the SAM-Encoder to predictive logits. For the second issue, we employ a logits-level adversarial discriminator to optimize the fine-tuning decoder and a feature-level adversarial prompted segmentor to provide prompted masks, enabling SAM to generate class-agnostic maps. Both the logits-level adversarial discriminator and the prompted segmentor help bridge the domain gap between features and predictive logits. Finally, we freeze SAM and jointly optimize the prompted segmentor and fine-tuning decoder in an end-to-end manner.
\par
\begin{figure}
\begin{center}
   \includegraphics[width=1.0\linewidth]{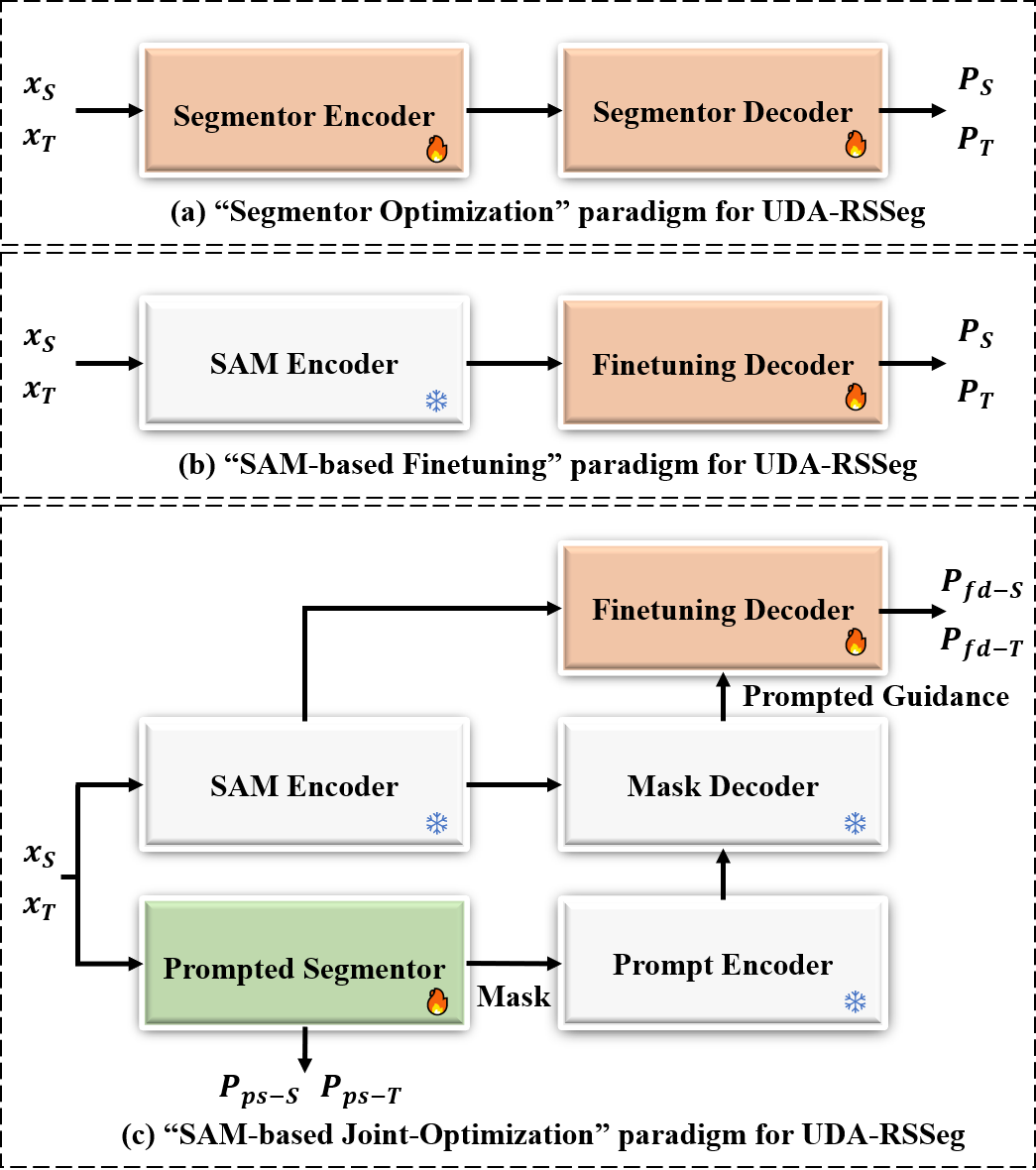}
\end{center}
   \caption{\textit{Paradigm comparison on UDA-RSSeg task.} \{$\bm{x_{S}},\bm{x_{T}}$\} and \{$\bm{P_{S}},\bm{P_{T}}$\} respectively denote the source/target images and predictions. \{$\bm{P_{S}^{FD}},\bm{P_{T}}^{FD}$\} and \{$\bm{P_{S}^{PS}},\bm{P_{T}^{PS}}$\} respectively denote the source/target predictions of finetuning decoder and prompted segmentor.}
\label{Fig1}
\end{figure}

 Fig.~\ref{Fig1} compares paradigms for the UDA-RSSeg task. Previous methods primarily use the ``Segmentor Optimization'' paradigm (Fig.~\ref{Fig1}(a)), where the encoder and decoder of the segmentor are optimized end-to-end using various learning strategies (e.g., adversarial learning, self-training). Fig.~\ref{Fig1}(b) illustrates the ``SAM-based Finetuning'' paradigm, which utilizes SAM to generate more generalized features for target images. By optimizing the fine-tuning decoder, these generalized features are mapped to predictive logits. In this paper, we introduce the "SAM-based Joint-Optimization" paradigm (Fig.~\ref{Fig1}(c)), where SAM provides generalized feature representation, and the prompted segmentor offers class-agnostic prompted guidance. By jointly optimizing the prompted segmentor and fine-tuning decoder, our paradigm significantly enhances segmentation performance on target images. Compared to ``Segmentor Optimization'' paradigm, our approach harnesses SAM's generalization capability to reduce feature inconsistency between source and target features. Compared to the "SAM-based Finetuning" paradigm, our method incorporates prompted guidance to address the domain gap between features and predictive logits.
\par We conduct extensive experiments on prominent benchmark datasets, including ISPRS (Potsdam/Vaihingen)\cite{ISPRS} and CITY-OSM (Paris/Chicago)\cite{CITY-OSM}. Comparative analyses demonstrate that SAM-JOANet outperforms previous state-of-the-art methods. Our visualization and analysis further highlight the interpretability of SAM-JOANet. The key contributions of this work are summarized as follows:
\begin{itemize}
\item We propose a joint-optimized adversarial learning framework based on the prompted foundation model (SAM). To the best of our knowledge, we are the first to introduce the ``SAM-based Joint-Optimization'' paradigm in the context of UDA-RSSeg.
\item We propose a prompted guidance mechanism that integrates the optimization of a prompted segmentor with a fine-tuning decoder, which effectively bridges the gap between class-agnostic maps and class-aware predictions.
\item  Our proposed SAM-JOANet achieves overall better segmentation results than previous SOTA methods on multiple prominent benchmark datasets.
\end{itemize}

\section{Related Work}
\label{sec2}
\subsection{Remote Sensing Semantic Segmentation}
\par Semantic segmentation task aims to categorize target objects in pixel-level. Fully convolutional networks (FCN)~\cite{FCN} is a pioneer deep learning based method on this task. Followed FCN, many notable methods~\cite{DeeplabV3, PSPNet, BiSeNetV2, SegFormer} are proposed, which significantly promote the development of this task.
\par On remote sensing scene, semantic segmentation task (RSSeg) is widely applied on geographical element analysis, urban/rural planning, disaster assessment etc. Semantic segmentation on remote sensing scene mainly faces the challenge from complex landscapes on large geographical region and large intra-class variance by different grounding sampling distance. To address these issues, many notable methods are proposed. Some methods~\cite{RSSeg-AMDF, RSSeg-DMSeg, GeoAgent} utilize the abundant information from multiple hierarchy features to achieve strong segmentation performance. DPFANet~\cite{DPFANet} and BSNet~\cite{RSSeg-BSNet} integrate adaptive feature fusion network and edge optimization block to enhance the representation ability from local to global features. With the development of Transformer structure, many methods~\cite{SAT, STDSNet, DistiRSSeg, RayTrans} design effective Transformer-based networks to exploit self-attention information for RSSeg task. 
\subsection{Unsupervised Domain Adaptation Semantic Segmentation for Remote Sensing}
\par Unsupervised domain adaptation (UDA) aims to adapt the knowledge of source-trained model to target samples. For unsupervised domain adaptation semantic segmentation on remote sensing scene (UDA-RSSeg), many excellent works are proposed in recent years. 
\par UDA-GAN~\cite{DA-UDAGAN} first introduces a GAN-based segmentation network to tackle UDA-RSSeg task. Since then, many methods~\cite{Maetal-MBATA, Wangetal-FGUDA, DNT, JDAF} have been developed that employ image-level and feature-level adversarial learning to align features between source and target images. Although adversarial learning can mitigate domain shift issues, it often fails to prevent the tendency of source-trained models to favor source image characteristics. To address this limitation, the focus has shifted toward non-adversarial paradigms using self-training mechanisms. Self-training methods~\cite{ST-UDA1, ST-DASegNet, ST-UDA2, Crots} typically rely on the exponential moving average (EMA) technique to generate pseudo-labels for target images. By training on these pseudo-labels, source-trained models are better equipped to adapt to target domains.
\subsection{Prompted Foundation Models for Remote Sensing}
The introduction of the Segment Anything Model (SAM)~\cite{SAM} has revolutionized the field of semantic segmentation. Leveraging prompted learning for guidance, SAM-based networks demonstrate remarkable generalization across diverse scenarios. In remote sensing, several innovative methods have been developed using the SAM framework. MAF-SAM~\cite{MAF-SAM} harnesses SAM's generalized capabilities to effectively process multi-spectral images. SCD-SAM~\cite{SCD-SAM} utilizes SAM’s robust generalization to enhance performance in semantic change detection tasks. Additionally, SAM-EDA~\cite{SAM-EDA} integrates SAM into an unsupervised domain adaptation network for semantic segmentation under adverse conditions. However, the UDA-RSSeg task remains underexplored. This paper introduces a new paradigm that leverages the generalized capabilities of SAM to ease the domain-shift in UDA-RSSeg.
\begin{figure*}
\begin{center}
   \includegraphics[width=1.0\linewidth]{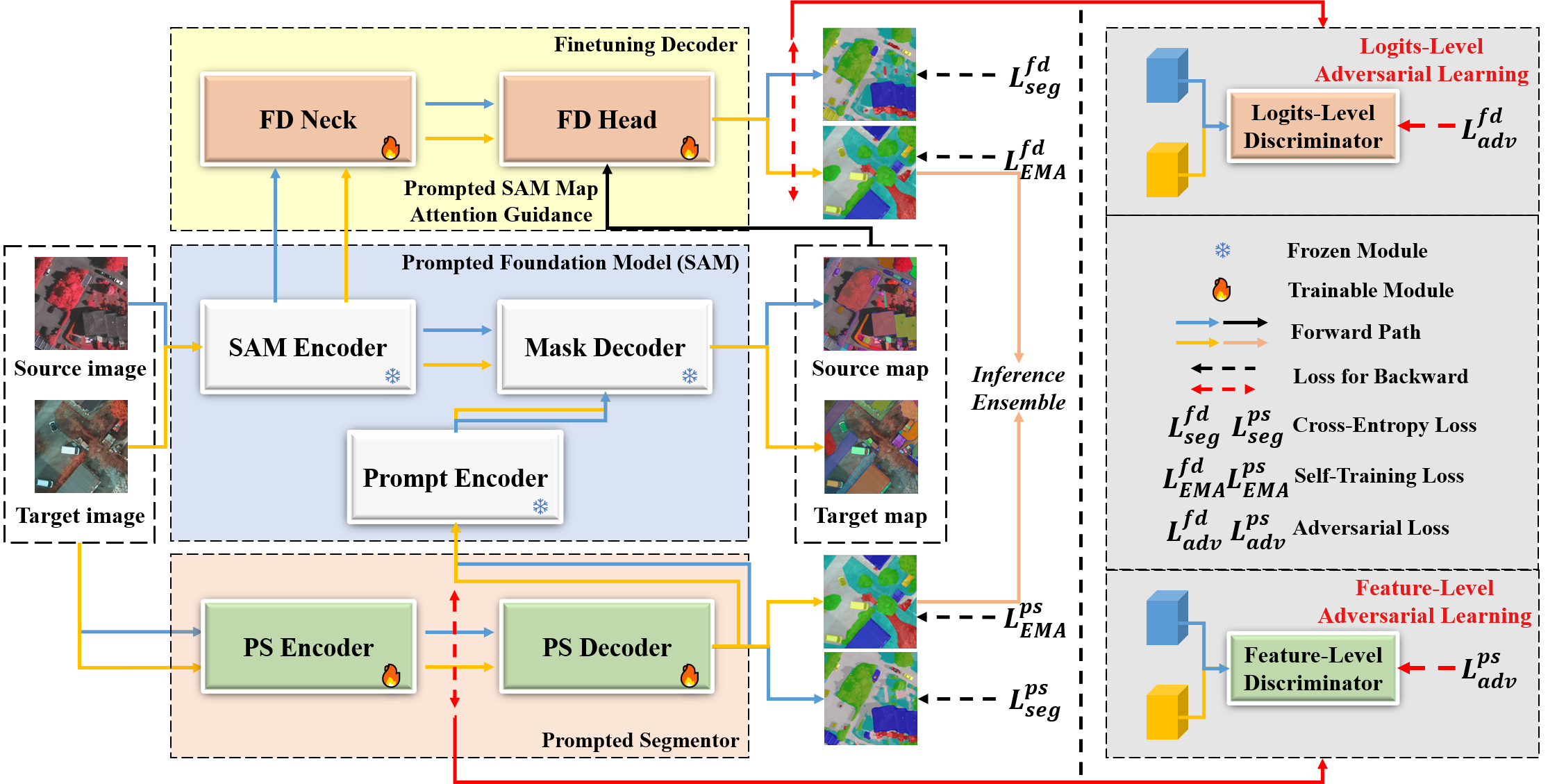}
\end{center}
   \caption{\textit{The overview of SAM-JOANet.} The architecture contains 3 modules, which are Prompted segmentor, Prompted foundation model (SAM) and Finetuning decoder. Adversarial learning and self-training are applied for joint-optimization.}
\label{Fig2}
\end{figure*}
\section{Proposed Method}
\subsection{Prompted Segmentor}
As shown in Fig.~\ref{Fig2}, prompted segmentor is designed to provide prompted mask for SAM to generate class-agnostic map. To generate prompted mask, both source and target images ({$\bm{x_{S}}, \bm{x_{T}}$}) are processed sequentially through the ``PS Encoder'' ($f_{ps-e}$) and ``PS Decoder'' ($f_{ps-d}$), which can be formulated in Eq.~\ref{Eq1} and Eq.~\ref{Eq2}.
\begin{equation}
\bm{F_{ps-S}} = f_{ps-e}(\bm{x_{S}}), \bm{F_{ps-T}} = f_{ps-e}(\bm{x_{T}})
\label{Eq1}
\end{equation}
\begin{equation}
\bm{P_{ps-S}} = f_{ps-d}(\bm{F_{ps-S}}), \bm{P_{ps-T}} = f_{ps-d}(\bm{F_{ps-T}})
\label{Eq2}
\end{equation}
where \{$\bm{F_{ps-S}},\bm{F_{ps-T}}$\} denote the feature maps of source and target images, respectively. \{$\bm{P_{ps-S}},\bm{P_{ps-T}}$\} denote the predictive logits of source and target images. 
\par The outputs {$\bm{P_{ps-S}}, \bm{P_{ps-T}}$} undergo a channel-wise $argmax$ operation, as detailed in Eq.~\ref{Eq3}, to generate the prompted masks {$\bm{P_{S}}, \bm{P_{T}}$}. These source and target prompted masks are then used as inputs for the prompted encoder.
\begin{equation}
\bm{{P}_{S}} = \mathop{\arg\max}\limits_{c}(\bm{P_{ps-S}}), \bm{{P}_{T}} = \mathop{\arg\max}\limits_{c}(\bm{P_{ps-T}})
\label{Eq3}
\end{equation}
\subsection{Prompted Foundation Model}
To consistently represent source and target images, we leverage the generalization capability of SAM. As shown in Fig.~\ref{Fig2}, ``SAM Encoder''($f_{sam-e}$) is used to extracted features for source and target images (Eq.~\ref{Eq4}). 
\begin{equation}
\bm{F_{s-S}} = f_{sam-e}(\bm{x_{S}}), \bm{F_{s-T}} = f_{sam-e}(\bm{x_{T}})
\label{Eq4}
\end{equation}
where $\bm{F_{s-S}}$ and $\bm{F_{s-T}}$ denote the output features from ``SAM Encoder''. These features will be served as input of finetuning decoder. Moreover, these features together with the features output from ``Prompt Encoder'' ($f_{sam-pe}$) are served as input of ``Mask Decoder'' ($f_{sam-md}$) to generate the class-agnostic maps (\{$\bm{M_{S}},\bm{M_{T}}$\}). This process can be formulated in Eq.~\ref{Eq5}.
\begin{equation}
\begin{cases}{}
   \bm{M_{S}} = f_{sam-md}(\bm{F_{s-S}}, f_{sam-pe}(\bm{P_{S}})) \\ \\ \bm{M_{T}} = f_{sam-md}(\bm{F_{s-T}}, f_{sam-pe}(\bm{P_{T}}))
\end{cases}
\label{Eq5}
\end{equation}
\begin{figure}
\begin{center}
   \includegraphics[width=1.0\linewidth]{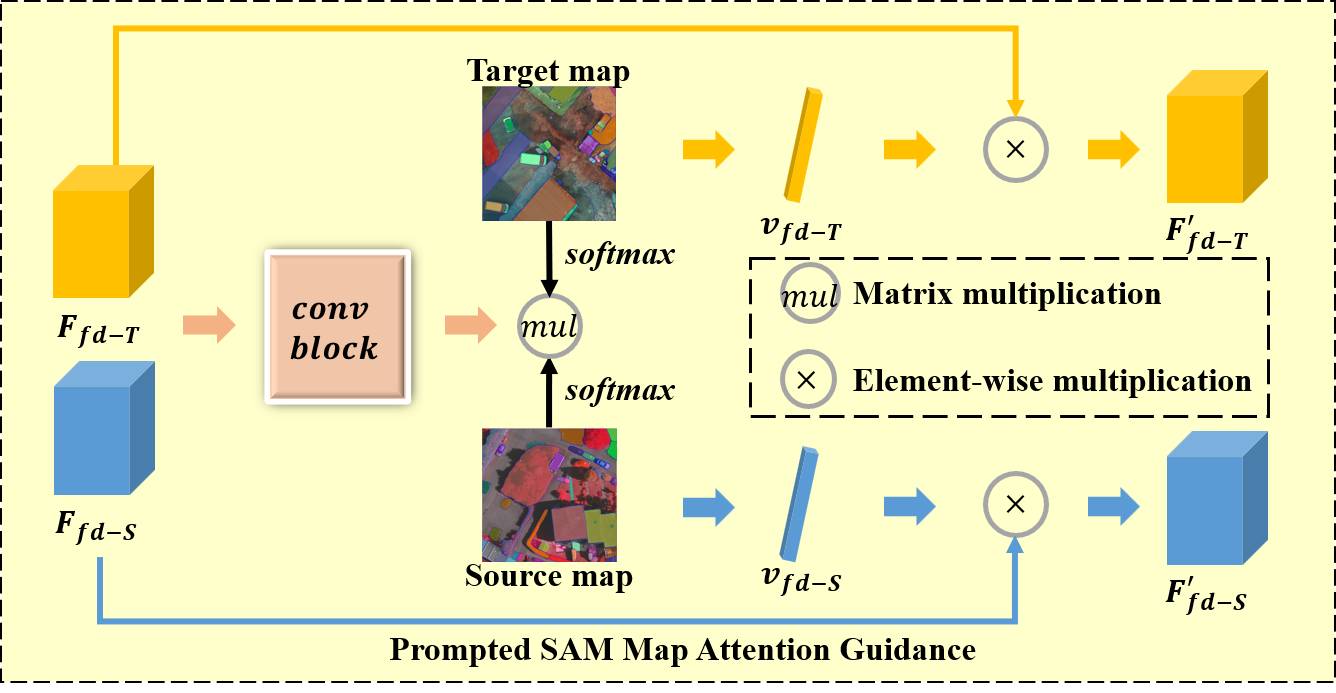}
\end{center}
   \caption{\textit{Prompted SAM map attention guidance block.} Here, ``conv block'' indicates [Conv-BN-ReLU] block.}
\label{Fig3}
\end{figure}
\subsection{Finetuning Decoder}
As shown in Fig.~\ref{Fig2}, finetuning decoder is designed to map the features of ``SAM Encoder'' to predictive logits with the guidance of SAM's class-agnostic maps. Finetuning decoder contains ``FD Neck'' ($f_{fd-n}$) and ``FD Head'' ($f_{fd-h}$). As shown in Eq.~\ref{Eq6}, output features from ``SAM Encoder'' (Eq.~\ref{Eq4}) first pass through ``FD Neck'' to map multi-scale features into single-scale features. 
\begin{equation}
\bm{F_{fd-S}} = f_{fd-n}(\bm{F_{s-S}}), \bm{F_{fd-T}} = f_{fd-n}(\bm{F_{s-T}})
\label{Eq6}
\end{equation}
where \{$\bm{F_{fd-S}},\bm{F_{fd-T}} \in \mathbb{R}^{C \times H \times W}$\} denote the output features from ``FD Neck''. 
\par Then, these features are guided by class-agnostic maps through attention mechanism. As shown in Fig.~\ref{Fig3}, features are fed into a ``conv block'' and conducted matrix multiplication with \{$\bm{M_{S}},\bm{M_{T}} \in \mathbb{R}^{H \times W}$\} to generate channel-weighted vectors (\{$\bm{v_{fd-S}},\bm{v_{fd-T}} \in \mathbb{R}^{C}$\}). This process is shown in Eq.~\ref{Eq7} and Eq.~\ref{Eq8}.
\begin{equation}
   v_{fd-S}^{c} = \frac{exp(\sum_{h, w=1}^{H, W}(f_{c}(\bm{F_{fd-S}}))^{c, h, w} \times \bm{M_{S}^{c, h, w}})}{\sum_{c=1}^{C}exp(\sum_{h, w=1}^{H, W}(f_{c}(\bm{F_{fd-S}}))^{c, h, w} \times \bm{M_{S}^{c, h, w}})}
\label{Eq7}
\end{equation}
\begin{equation}
   v_{fd-T}^{c} = \frac{exp(\sum_{h, w=1}^{H, W}(f_{c}(\bm{F_{fd-T}}))^{c, h, w} \times \bm{M_{T}^{c, h, w}})}{\sum_{c=1}^{C}exp(\sum_{h, w=1}^{H, W}(f_{c}(\bm{F_{fd-T}}))^{c, h, w} \times \bm{M_{T}^{c, h, w}})}
\label{Eq8}
\end{equation}
where $\bm{v_{fd-S}} = [v_{fd-S}^{1}, \cdots, v_{fd-S}^{C}]$, $\bm{v_{fd-T}} = [v_{fd-T}^{1}, \cdots, v_{fd-T}^{C}]$. These vectors set the ratio of features' channel, which can guide the features in a channel-selection manner (Eq.~\ref{Eq9}). 
\begin{equation}
   \bm{F_{fd-S}^{'}} = \bm{F_{fd-S}} \times \bm{v_{fd-S}}, \bm{F_{fd-T}^{'}} = \bm{F_{fd-T}} \times \bm{v_{fd-T}}
\label{Eq9}
\end{equation}
Finally, attention guided features (\{$\bm{F_{fd-S}^{'}},\bm{F_{fd-T}^{'}} \in \mathbb{R}^{C \times H \times W}$\}) are mapped into predictive logits (\{$\bm{P_{fd-S}},\bm{P_{fd-T}}$\}) by ``FD Head'', which is formulated as Eq.~\ref{Eq10}.   
\begin{equation}
   \bm{P_{fd-S}} = f_{fd-h}(\bm{F_{fd-S}^{'}}), \bm{P_{fd-T}} = f_{fd-h}(\bm{F_{fd-T}^{'}})
\label{Eq10}
\end{equation}
\subsection{Joint-Optimized Mechanism with Combined Loss}
To jointly optimize SAM-JOANet, we integrate adversarial learning and self-training strategy to build connections between generalized feature and predictive logits.  
\subsubsection{Optimizing Finetuning Decoder.} To optimize finetuning decoder, we involve logits-level adversarial learning strategy to enhance the consistent feature representation of ``FD Neck'' and ``FD-Head''. Towards source and target predictive logits, we design logits-level discriminator ($D_{l}$) for alignment. The adversarial loss ($\mathcal{L}_{adv}^{fd}$) is calculated in Eq.~\ref{Eq11}. Here, $\theta_{fd}$ and $\theta_{dl}$ denote the trainable parameters of finetuning decoder and logits-level discriminator, respectively. 
\begin{equation}
\begin{split}
   \mathcal{L}_{adv}^{fd}(\theta_{fd}, \theta_{dl}) &= \mathbb{E}_{x^{s} \sim X^{s}}[log(D_{l}(\bm{P_{fd-S}}))]  \\
                  &+ \mathbb{E}_{x^{t} \sim X^{t}}[log(1 - D_{l}(\bm{P_{fd-T}}))]
   \end{split}
\label{Eq11}
\end{equation}
\par To further alleviate the representation tendency on source annotated samples, we embed self-training mechanism into SAM-JOANet to enhance optimization effect on finetuning decoder. The first step is to update ``EMA-Finetuning Decoder'' ($f_{ema-fd}$) by exponential moving average technique. This process can be formulated in Eq.~\ref{Eq12}. 
\begin{equation}
\theta_{ema-fd}^{t} = \alpha\theta_{ema-fd}^{t-1} + (1-\alpha)\theta_{fd}^{t}
\label{Eq12}
\end{equation}
\par where $\alpha$ denotes the decay factors controlling the updating rate. $\theta_{ema-fd}^{t}$ refers to trainable parameters of ``EMA-Finetuning Decoder'' (including ``EMA-FD Neck'' and ``EMA-FD Head'') at $t^{th}$ step.
\par The second step is to generate pseudo-labels ($\bm{\hat{P}_{fd}}$) for target image, which is shown in Eq.~\ref{Eq13}.
\begin{equation}
\bm{\hat{P}_{fd}} = \mathop{\arg\max}\limits_{c}f_{ema-fd}(\bm{F_{s-T}}, \bm{M_{T}})
\label{Eq13}
\end{equation}
\par The third step is to construct self-training loss function to optimize finetuning decoder, which is formulated in Eq.~\ref{Eq14}.
\begin{equation}
\begin{split}
    \mathcal{L}_{EMA}^{fd}(\theta_{fd}) = -\sum_{h, w=1}^{H, W}\sum_{c=1}^{C}\bm{\hat{P}_{fd}^{(h, w, c)}}log(\bm{P_{fd-T}^{(h, w, c)}}) 
\end{split}
    \label{Eq14}
\end{equation}
\subsubsection{Optimizing Prompted Segmentor.} A high-performance prompted segmentor is crucial for providing a high-quality prompted mask and thus, enhancing its performance is also important. To achieve this, we implement feature-level adversarial learning to align the features of the ``PS Encoder'' shown in Eq.~\ref{Eq15}. Given that ``PS Encoder'' lacks generalized feature representation capabilities, it is essential to improve the feature consistency between source and target images.
\begin{equation}
\begin{split}
   \mathcal{L}_{adv}^{ps}(\theta_{ps-e}, \theta_{df}) &= \mathbb{E}_{x^{s} \sim X^{s}}[log(D_{f}(\bm{F_{ps-S}}))]  \\
                  &+ \mathbb{E}_{x^{t} \sim X^{t}}[log(1 - D_{f}(\bm{F_{ps-T}}))]
   \end{split}
\label{Eq15}
\end{equation}
where $\mathcal{L}_{adv}^{ps}$ denotes the adversarial loss. $\theta_{ps-e}$ and $\theta_{df}$ respectively denote the trainable parameters of ``PS Encoder'' and feature-level discriminator ($D_{f}$). 
\par Similar to the optimization of finetuning decoder (Eq.~\ref{Eq13}, Eq.~\ref{Eq14}), we also apply self-training strategy on prompted segmentor. The ``EMA Prompted Segmentor'' ($f_{ema-ps}$) updating, pseudo-label generation and loss function construction are respectively shown in Eq.~\ref{Eq16}, Eq.~\ref{Eq17} and Eq.~\ref{Eq18}. 
\begin{equation}
\theta_{ema-ps}^{t} = \alpha\theta_{ema-ps}^{t-1} + (1-\alpha)\theta_{ps}^{t}
\label{Eq16}
\end{equation}
\begin{equation}
\bm{\hat{P}_{ps}} = \mathop{\arg\max}\limits_{c}f_{ema-ps}(\bm{x_{T}})
\label{Eq17}
\end{equation}
\begin{equation}
\begin{split}
    \mathcal{L}_{EMA}^{ps}(\theta_{ps}) = -\sum_{h, w=1}^{H, W}\sum_{c=1}^{C}\bm{\hat{P}_{ps}^{(h, w, c)}}log(\bm{P_{ps-T}^{(h, w, c)}}) 
\end{split}
    \label{Eq18}
\end{equation}
where $\theta_{ema-ps}$ and $\theta_{ps}$ denote the trainable parameters of ``EMA-Prompted Segmentor'' and ``Prompted Segmentor'', respectively. $\bm{\hat{P}_{ps}}$ denotes the pseudo-label of target image.
\begin{algorithm}[H]
\normalsize
		\caption{Joint-Optimized Paradigm on SAM-JOANet}
		\label{Alg1}
		\begin{algorithmic}
			\REQUIRE Source and target images, $\bm{x_{S}}$ and $\bm{x_{T}}$. Source-label $\bm{y_{S}}$. Trainable parameters, $\theta_{ps}$, $\theta_{fd}$. Frozen parameters of SAM-JOANet. The training iteration is set as $K$.
			\ENSURE {Updated trainable parameters, $\theta_{ps}^{'}$, $\theta_{fd}^{'}$.}
			\FOR {$k = 1$, \ldots, $K$}
                \STATE {\textbf{$\bm{1^{st}}$ Stage Optimization on Prompted Segmentor:}} 
                \STATE {1:\quad Get feature maps, $\bm{F_{ps-S}}, \bm{F_{ps-T}}$ using Eq.~\ref{Eq1}.}
                \STATE {2:\quad Get predictive logits, $\bm{P_{ps-S}}, \bm{P_{ps-T}}$ using Eq.~\ref{Eq2}.}
                \STATE {3:\quad Get prompted masks, $\bm{P_{S}}, \bm{P_{T}}$ using Eq.~\ref{Eq3}.}
                \STATE {4:\quad Calculate the segmentation cross-entropy loss, $L_{seg}^{ps}$ with source-label.}
                \STATE {5:\quad Calculate the feature-level adversarial loss, $L_{adv}^{ps}$ using Eq.~\ref{Eq15}. Use ``$min-max$'' criterion to optimize.}
                \STATE {6:\quad Calculate the self-training loss, $L_{EMA}^{ps}$ using Eq.~\ref{Eq16} $\sim$ Eq.~\ref{Eq18}.}
                \STATE {7:\quad Compute gradients by backward operation with combined loss, $L^{ps} = L_{seg}^{ps} + \gamma_{1} L_{adv}^{ps} + \gamma_{2} L_{EMA}^{ps}$.}
                \STATE {8:\quad Update trainable parameters, $\theta_{ps}$ by $step$.}
                \STATE {9:\quad Frozen prompted segmentor.}
                \STATE {\textbf{$\bm{2^{nd}}$ Stage Optimization on Finetuning Decoder:}} 
                \STATE {10:\quad Get generalized feature maps from SAM, $\bm{F_{s-S}}, \bm{F_{s-T}}$ using Eq.~\ref{Eq4}.}
                \STATE {11:\quad Get class-agnostic maps, $\bm{M_{S}}, \bm{M_{T}}$ using Eq.~\ref{Eq5}.}
                \STATE {12:\quad Get feature maps, $\bm{F_{fd-S}}, \bm{F_{fd-T}}$ using Eq.~\ref{Eq6}.}
                \STATE {13:\quad Get attention guided feature maps, $\bm{F_{fd-S}^{'}}, \bm{F_{fd-T}^{'}}$ using Eq.~\ref{Eq7} $\sim$ Eq.~\ref{Eq9}.}
                \STATE {14:\quad Get predictions, $\bm{P_{fd-S}}, \bm{P_{fd-T}}$ using Eq.~\ref{Eq10}.}
                \STATE {15:\quad Calculate the segmentation cross-entropy loss, $L_{seg}^{fd}$ with source-label.}
                \STATE {16:\quad Calculate the logits-level adversarial loss, $L_{adv}^{fd}$ using Eq.~\ref{Eq11}. Use ``$min-max$'' criterion to optimize.}
                \STATE {17:\quad Calculate the self-training loss, $L_{EMA}^{fd}$ using Eq.~\ref{Eq12} $\sim$ Eq.~\ref{Eq14}.}
                \STATE {18:\quad Compute gradients by backward operation with combined loss, $L^{fd} = L_{seg}^{fd} + \gamma_{3} L_{adv}^{fd} + \gamma_{4} L_{EMA}^{fd}$.}
                \STATE {19:\quad Update trainable parameters, $\theta_{fd}$ by $step$.}
                \STATE {20:\quad Frozen Finetuning Decoder.}
            \ENDFOR
		\end{algorithmic}
\end{algorithm}
\subsubsection{Two-Stage Jointly Optimization with Combined Loss.}
To optimize SAM-JOANet, we propose a two-stage jointly optimizing paradigm. In a single iteration, the first stage focuses on optimizing the prompted segmentor, where trainable parameters are updated during the backward pass. The second stage involves optimizing the fine-tuning decoder. Additionally, our proposed two-stage optimization paradigm operates as an end-to-end learning system. The whole optimization process is shown in Alg.~\ref{Alg1}.
\section{Experiments and Analysis}
\subsection{Datasets and Metrics}
\subsubsection{Datasets} To evaluate SAM-JOANet on UDA-RSSeg task, we select two benchmark datasets:ISPRS~\cite{ISPRS} and CITY-OSM~\cite{CITY-OSM}.
\par ISPRS~\cite{ISPRS} provides a rich collection of image samples for diverse tasks within the field of remote sensing. Each image in this dataset is meticulously annotated at the pixel level across six categories: ``Clutter,'' ``Impervious Surfaces,'' ``Car,'' ``Tree,'' ``Low Vegetation,'' and ``Building''. ISPRS contains two main subsets: Potsdam and Vaihingen. The Potsdam dataset consists of 38 very-high-resolution true orthophotos (VHR TOPs), each with dimensions of 6000 $\times$ 6000 pixels. This dataset includes images in three imaging modes: IR-R-G, R-G-B, and R-G-B-IR. The IR-R-G and R-G-B images are composed of three channels, whereas the R-G-B-IR images include four channels. For our experiments, we focus on the IR-R-G and R-G-B images. The Vaihingen dataset comprises 33 VHR TOPs, approximately sized at 2000 $\times$ 2000 pixels, and exclusively utilizes the IR-R-G imaging mode. 
\par To ensure a fair comparison with previous methodologies~\cite{DA-DualGAN, DA-CLAL, DA-CSLG, ST-DASegNet}, we follow their data preprocessing technique, which involves cropping very high-resolution (VHR) images into smaller patches. These patches are standardized at a size of 512 $\times$ 512 pixels. For the cropping process, we apply strides of 512 for the Potsdam dataset and 256 for the Vaihingen dataset, resulting in totals of 4598 and 1696 patches, respectively. Additionally, we utilize the same strategy for dataset division, separating each into training and testing subsets. Consequently, the training subsets for Potsdam and Vaihingen include 2904 and 1296 images, respectively, while the testing subsets comprise 1694 and 440 images, respectively. 
\par In this paper, we design four UDA-RSSeg tasks, which are listed as follows.
\begin{itemize}
\item Adapt Potsdam IR-R-G to Vaihingen IR-R-G (Potsdam IR-R-G $\rightarrow$ Vaihingen IR-R-G).
\item Adapt Vaihingen IR-R-G to Potsdam IR-R-G (Vaihingen IR-R-G $\rightarrow$ Potsdam IR-R-G).
\item Adapt Potsdam R-G-B to Vaihingen IR-R-G (Potsdam R-G-B $\rightarrow$ Vaihingen IR-R-G).
\item Adapt Vaihingen IR-R-G to Potsdam R-G-B (Vaihingen IR-R-G $\rightarrow$ Potsdam R-G-B).
\end{itemize}
\par CITY-OSM focuses exclusively on urban areas, capturing the intricate details of cities, including streets, buildings, parks, and other urban features. CITY-OSM~\cite{CITY-OSM} comprises several sub-sets, Berlin, Chicago, Zurich, Paris and Tokyo. All images are annotated in pixel-level with 3 categories including ``Background'', ``Road'' and ``Building''. Following previous methods~\cite{DA-UDA1, ST-DASegNet}, we select Paris and Chicago sub-sets to conduct experiments on UDA-RSSeg task. Paris and Chicago datasets respectively have 725 and 457 images. Similar to the data-preprocessing approach on ISPRS, we also adopt cropping on CITY-OSM, where the stride and patch size are respectively set as 512 and 512 $\times$ 512. After cropping, Paris and Chicago datasets respectively have 22500 and 13710 images, where 70\% are randomly selected as training images and the remaining 30\% are selected as testing images.
\par In this paper, we follow~\cite{DA-UDA1, ST-DASegNet} and conduct one UDA-RSSeg task, which is listed as follows.
\begin{itemize}
\item Adapt Paris to Chicago (Paris $\rightarrow$ Chicago).
\end{itemize}
\begin{table*}	
	\centering
    \begin{adjustbox}{scale=0.75}
    \begin{tabular}{ccccccccccccccc}
		\cmidrule(r){1-15}
  \multirow{2}{*}{Methods} &  \multicolumn{2}{c}{Clutter} & \multicolumn{2}{c}{Impervious surfaces} & \multicolumn{2}{c}{Car} & \multicolumn{2}{c}{Tree} & \multicolumn{2}{c}{Low vegetation} & \multicolumn{2}{c}{Building} & \multicolumn{2}{c}{Overall}
  \\ \cmidrule(r){2-15}
  {} & $IoU$ & $F1$-score & $IoU$ & $F1$-score & $IoU$ & $F1$-score & $IoU$ & $F1$-score & $IoU$ & $F1$-score & $IoU$ & $F1$-score & $mIoU$ & $mF$-score 
  \\ \cmidrule(r){1-15}
  AdaptSegNet~\cite{AdapSegNet} & 4.60 & 8.76 & 54.39 & 70.39 & 6.40 & 11.99 & 52.65 & 68.96 & 28.98 & 44.91 & 63.14 & 77.40 & 35.02 & 47.05 
  \\
  ProDA~\cite{ProDA} & 3.99 & 8.21 & 62.51 & 76.85 & 39.20 & 56.52 & 56.26 & 72.09 & 34.49 & 51.65 & 71.61 & 82.95 & 44.68 & 58.05
  \\
  DualGAN~\cite{DA-DualGAN} & 29.66 & 45.65 & 49.41 & 66.13 & 34.34 & 51.09 & 57.66 & 73.14 & 38.87 & 55.97 & 62.30 & 76.77 & 45.38 & 61.43 
  \\
  Zhang~\textit{et al.}~\cite{DA-CSLG} & 20.71 & 31.34 & 67.74 & 80.13 & 44.90 & 61.94 & 55.03 & 71.90 & 47.02 & 64.16 & 76.75 & 86.65 & 52.03 & 66.02
  \\
  Wang~\textit{et al.}~\cite{DA-FGUDA} & 21.85 & 35.87 & \textbf{76.58} & \textbf{86.73} & 35.44 & 52.33 & 55.22 & 71.15 & 49.97 & 66.64 & 82.74 & 90.56 & 53.63 & 67.21 
  \\
  DNT~\cite{DNT} & 14.77 & 25.74 & 69.74 & 82.18 & \textbf{53.88} & \textbf{70.03} & 59.19 & 74.37 & 47.51 & 64.42 & 80.04 & 88.91 & 54.19 & 67.61
  \\
  CIA-UDA~\cite{DA-CIAUDA} & 27.80 & 43.51 & 63.28 & 77.51 & 52.91 & 69.21 & 64.11 & 78.13 & 48.03 & 64.90 & 75.13 & 85.80 & 55.21 & 69.84
  \\ 
  JDAF~\cite{JDAF} & 38.65 & 55.75 & 68.76 & 81.49 & 42.76 & 59.90 & 58.38 & 73.72 & 47.39 & 64.30 & 77.19 & 87.13 & 55.52 & 70.38
  \\
  DAFormer~\cite{DAFormer} & 48.26 & 60.17 & 74.09 & 84.12 & 38.96 & 56.41 & \textbf{70.88} & 81.36 & \textbf{57.53} & \textbf{71.48} & 84.07 & 90.75 & 62.30 & 74.05
  \\
  ST-DASegNet~\cite{ST-DASegNet} & 67.03 & 80.28 & 74.43 & 85.36 & 43.38 & 60.49 & 67.36 & 80.49 & 48.57 & 65.37 & \textbf{85.23} & \textbf{92.03} & 64.33 & 77.34
  \\ \cmidrule(r){1-15}
  \rowcolor{gray!30!} SAM-JOANet (ours) & \textbf{68.88} & \textbf{81.57} & 72.16 & 83.83 & 52.68 & 69.00 & 69.23 & \textbf{81.82} & 54.39 & 70.46 & 83.81 & 91.19 & \textbf{66.86} & \textbf{79.65}
  \\ \cmidrule(r){1-15}
   \end{tabular}
   \end{adjustbox}
   \caption{UDA-RSSeg comparison results (\%) on ``Potsdam IR-R-G $\rightarrow$ Vaihingen IR-R-G'' task.}
 \label{Tab1}
 \end{table*}
 \begin{table*}	
	\centering
    \begin{adjustbox}{scale=0.75}
    \begin{tabular}{ccccccccccccccc}
		\cmidrule(r){1-15}
  \multirow{2}{*}{Methods} &  \multicolumn{2}{c}{Clutter} & \multicolumn{2}{c}{Impervious surfaces} & \multicolumn{2}{c}{Car} & \multicolumn{2}{c}{Tree} & \multicolumn{2}{c}{Low vegetation} & \multicolumn{2}{c}{Building} & \multicolumn{2}{c}{Overall}
  \\ \cmidrule(r){2-15}
  {} & $IoU$ & $F1$-score & $IoU$ & $F1$-score & $IoU$ & $F1$-score & $IoU$ & $F1$-score & $IoU$ & $F1$-score & $IoU$ & $F1$-score & $mIoU$ & $mF$-score 
  \\ \cmidrule(r){1-15}
  AdaptSegNet~\cite{AdapSegNet} & 8.36 & 15.33 & 49.55 & 64.64 & 40.95 & 58.11 & 22.59 & 36.79 & 34.43 & 61.50 & 48.01 & 63.41 & 33.98 & 49.96
  \\
  ProDA~\cite{ProDA} & 10.63 & 19.21 & 44.70 & 61.72 & 46.78 & 63.74 & 31.59 & 48.02 & 40.55 & 57.71 & 56.85 & 72.49 & 38.51 & 53.82
  \\
  DualGAN~\cite{DA-DualGAN} & 11.48 & 20.56 & 51.01 & 67.53 & 48.49 & 65.31 & 34.98 & 51.82 & 36.50 & 53.48 & 53.37 & 69.59 & 39.30 & 54.71
  \\
  DNT~\cite{DNT} & 11.51 & 20.65 & 61.91 & 76.48 & 49.50 & 66.22 & 35.46 & 52.36 & 37.61 & 54.67 & 66.41 & 79.81 & 43.74 & 58.36
  \\
  Zhang~\textit{et al.}~\cite{DA-CSLG} & 12.31 & \textbf{24.59} & 64.39 & 78.59 & 59.35 & 75.08 & 37.55 & 54.60 & 47.17 & 63.27 & 66.44 & 79.84 & 47.87 & 62.66 
  \\
  Wang~\textit{et al.}~\cite{DA-FGUDA} & 11.65 & 19.47 & 73.43 & 84.55 & 63.86 & 77.85 & 32.68 & 47.36 & 47.69 & 63.45 & 76.32 & 87.43 & 50.94 & 63.31 
  \\
  JDAF~\cite{JDAF} & \textbf{13.10} & 23.17 & 67.70 & 80.74 & 63.22 & 77.47 & 36.21 & 53.17 & 51.19 & 67.72 & 76.36 & 86.59 & 51.30 & 64.81
  \\
  CIA-UDA~\cite{DA-CIAUDA} & 10.87 & 19.61 & 62.74 & 77.11 & 65.35 & 79.04 & 47.74 & 64.63 & 54.40 & 70.47 & 72.31 & 83.93 & 52.23 & 65.80
  \\ 
  DAFormer~\cite{DAFormer} & 2.56 & 5.02 & 68.42 & 79.07 & 65.20 & 79.31 & \textbf{70.65} & \textbf{82.13} & 56.39 & 72.48 & 78.94 & 87.64 & 57.03 & 67.61
  \\
  ST-DASegNet~\cite{ST-DASegNet} & 0.18 & 0.35 & \textbf{76.45} & \textbf{86.65} & 73.54 & 84.76 & 62.89 & 77.22 & \textbf{61.04} & \textbf{75.80} & 83.81 & \textbf{91.19} & \textbf{59.65} & 69.33
  \\ \cmidrule(r){1-15}
  \rowcolor{gray!30!} SAM-JOANet (ours) & 5.72 & 8.44 & 75.97 & 85.78 & \textbf{74.24} & \textbf{85.35} & 68.62 & 80.91 & 58.35 & 75.61 & \textbf{84.37} & 90.18 & \textbf{61.21} & \textbf{71.05}
  \\ \cmidrule(r){1-15}
   \end{tabular}
   \end{adjustbox}
   \caption{UDA-RSSeg comparison results (\%) on ``Vaihingen IR-R-G $\rightarrow$ Potsdam IR-R-G'' task.}
 \label{Tab2}
 \end{table*}
\subsubsection{Metrics} To evaluate the model's performance, we adopt $IoU/mIoU$ and $F1$-score/$mF$-score as metrics. Specifically, for a class $i$, $IoU$ is formulated as $IoU_{i} = tp_{i} / (tp_{i} + fp_{i} + fn_{i})$, where $tp_{i}$, $fp_{i}$, $fn_{i}$ denote true positive, false positive and false negative, respectively. The $mIoU$ is the mean value of all categories' $IoU$. Additionally, $F1$-score is defined as $F1$-score $= (2 \times Precision \times Recall) / (Precision + Recall)$. The $mF$-score is the mean value of all categories' $F1$-score.
\subsection{Implementation Details}
\subsubsection{Architecture Details.} 
SAM-JOANet consists of 3 key modules, which are prompted segmentor, prompted foundation model (SAM) and finetuning decoder. For prompted segmentor, we select SegFormer-b5~\cite{SegFormer}. ``PS Encoder'' and ``PS Decoder'' indicate ``mit-b5'' and ``ALL-MLP'', respectively. For SAM, there are 3 main architecture types, ``base'', ``large'' and ``huge''. Each type indicates a specific ViT-based ~\cite{ViT} ``SAM Encoder''. In this paper, we select ``base'' type for efficient training. For finetuning decoder, we select multi-level neck as ``FD Neck'' and select UperNet~\cite{UperNet} as ``FD Head''. For the logits-level and feature-level discriminators, we select PatchGAN~\cite{PatchGAN} to conduct adversarial learning. Specifically, two discriminators both have 4 convolutional blocks with kernels as size of 4 $\times$ 4. The stride settings are 2 for the first two blocks and 1 for the last two. The output channels for these blocks are set at 64, 128, 256, and 1, respectively. For logits-level discriminator and feature-level discriminator, the input-channel is equal with the category's number and the output channel of ``PS Encoder'', respectively. 
\subsubsection{Optimization Details} 
To implement the joint-optimized mechanism in SAM-JOANet, we have developed separate multi-optimizers for each critical component. For the ``Prompted Segmentor'' and ``Finetuning Decoder'', we employ AdamW~\cite{AdamW} as the optimizer. The initial learning rate is set at 0.00006, with a weight decay of 0.01. For the two discriminators, we use Adam~\cite{Adam} as the optimizer, with an initial learning rate of 0.0001 and the same weight decay of 0.01.
\subsubsection{Experimental Environment} 
All experiments are conducted using the mmsegmentation framework~\cite{mmseg2020}, a leading open-source platform designed to advance research and development in semantic segmentation. Built on PyTorch, mmsegmentation offers a comprehensive suite of cutting-edge algorithms, pre-trained models, and robust tools for researchers. For implementing adversarial learning and self-training within SAM-JOANet, we also utilize resources from Mmagic~\cite{MMagic}. All experiments are performed on two NVIDIA RTX 4090 GPUs, with a batch size of 3 per GPU.
\par For more detailed implementation details, please refer to \url{https://github.com/CV-ShuchangLyu/SAM-JOANet/}.
\begin{table*}	
	\centering
    \begin{adjustbox}{scale=0.75}
    \begin{tabular}{ccccccccccccccc}
		\cmidrule(r){1-15}
  \multirow{2}{*}{Methods} &  \multicolumn{2}{c}{Clutter} & \multicolumn{2}{c}{Impervious surfaces} & \multicolumn{2}{c}{Car} & \multicolumn{2}{c}{Tree} & \multicolumn{2}{c}{Low vegetation} & \multicolumn{2}{c}{Building} & \multicolumn{2}{c}{Overall}
  \\ \cmidrule(r){2-15}
  {} & $IoU$ & $F1$-score & $IoU$ & $F1$-score & $IoU$ & $F1$-score & $IoU$ & $F1$-score & $IoU$ & $F1$-score & $IoU$ & $F1$-score & $mIoU$ & $mF$-score 
  \\ \cmidrule(r){1-15}
  AdaptSegNet~\cite{AdapSegNet} & 2.99 & 5.81 & 51.26 & 67.77 & 10.25 & 18.54 & 51.51 & 68.02 & 12.75 & 22.61 & 60.72 & 75.55 & 31.58 & 43.05 
  \\
  ProDA~\cite{ProDA} & 2.39 & 5.09 & 49.04 & 66.11 & 31.56 & 48.16 & 49.11 & 65.86 & 32.44 & 49.06 & 68.94 & 81.89 & 38.91 & 52.70
  \\
  DualGAN~\cite{DA-DualGAN} & 3.94 & 13.88 & 49.16 & 61.33 & 40.31 & 57.88 & 55.82 & 70.66 & 27.85 & 42.17 & 65.44 & 83.00 & 39.93 & 54.82
  \\
  Zhang~\textit{et al.}~\cite{DA-CSLG} & 12.38 & 21.55 & 64.47 & 77.76 & 43.43 & 60.05 & 52.83 & 69.62 & 38.37 & 55.94 & 76.87 & 86.95 & 48.06 & 61.98 
  \\
  Wang~\textit{et al.}~\cite{DA-FGUDA} & 12.61 & 22.39 & 73.80 & 84.92 & 43.24 & 60.38 & 44.41 & 61.50 & 43.27 & 60.40 & 83.76 & 91.16 & 50.18 & 63.46 
  \\
  CIA-UDA~\cite{DA-CIAUDA} & 13.50 & 23.78 & 62.63 & 77.02 & 52.28 & 68.66 & 63.43 & 77.62 & 33.31 & 49.97 & 79.71 & 88.71 & 50.81 & 64.29
  \\ 
  JDAF~\cite{JDAF} & 32.71 & 49.30 & 64.33 & 78.29 & 45.87 & 62.90 & 51.99 & 68.41 & 42.16 & 59.31 & 75.53 & 86.06 & 52.10 & 67.38
  \\
  DNT~\cite{DNT} & 11.55 & 20.71 & 67.94 & 80.91 & \textbf{52.64} & \textbf{68.97} & 58.43 & 73.76 & \textbf{43.63} & \textbf{61.05} & 81.09 & 89.56 & 52.60 & 65.83
  \\
  DAFormer~\cite{DAFormer} & 22.57 & 33.72 & 67.44 & 79.65 & 45.60 & 60.13 & 66.27 & \textbf{80.41} & 40.49 & 54.93 & 81.34 & 90.07 & 53.95 & 66.49
  \\
  ST-DASegNet~\cite{ST-DASegNet} & \textbf{36.03} & \textbf{50.64} & 68.36 & \textbf{81.28} & 43.15 & 60.28 & 64.65 & 78.31 & 34.69 & 47.08 & \textbf{84.09} & \textbf{91.33} & 55.16 & 68.15
  \\ \cmidrule(r){1-15}
  \rowcolor{gray!30!} SAM-JOANet (ours) & 32.56 & 47.83 & \textbf{70.05} & 81.16 & 49.61 & 65.88 & \textbf{66.45} & 79.50 & 39.76 & 57.65 & 81.01 & 89.44 & \textbf{56.61} & \textbf{70.21}
  \\ \cmidrule(r){1-15}
   \end{tabular}
   \end{adjustbox}
   \caption{UDA-RSSeg comparison results (\%) on ``Potsdam R-G-B $\rightarrow$ Vaihingen IR-R-G''.}
 \label{Tab3}
 \end{table*}
 \begin{table*}	
	\centering
    \begin{adjustbox}{scale=0.75}
    \begin{tabular}{ccccccccccccccc}
		\cmidrule(r){1-15}
  \multirow{2}{*}{Methods} &  \multicolumn{2}{c}{Clutter} & \multicolumn{2}{c}{Impervious surfaces} & \multicolumn{2}{c}{Car} & \multicolumn{2}{c}{Tree} & \multicolumn{2}{c}{Low vegetation} & \multicolumn{2}{c}{Building} & \multicolumn{2}{c}{Overall}
  \\ \cmidrule(r){2-15}
  {} & $IoU$ & $F1$-score & $IoU$ & $F1$-score & $IoU$ & $F1$-score & $IoU$ & $F1$-score & $IoU$ & $F1$-score & $IoU$ & $F1$-score & $mIoU$ & $mF$-score 
  \\ \cmidrule(r){1-15}
  AdaptSegNet~\cite{AdapSegNet} & 6.11 & 11.50 & 37.66 & 59.55 & 42.31 & 55.95 & 30.71 & 45.51 & 15.10 & 25.81 & 54.25 & 70.31 & 31.02 & 44.75
  \\
  ProDA~\cite{ProDA} & 11.13 & 20.51 & 44.77 & 62.03 & 41.21 & 59.27 & 30.56 & 46.91 & 35.84 & 52.75 & 46.37 & 63.06 & 34.98 & 50.76
  \\
  DualGAN~\cite{DA-DualGAN} & 13.56 & 23.84 & 45.96 & 62.97 & 39.71 & 56.84 & 25.80 & 40.97 & 41.73 & 58.87 & 59.01 & 74.22 & 37.63 & 52.95
  \\
  DNT~\cite{DNT} & 8.43 & 15.55 & 56.41 & 72.13 & 46.78 & 63.74 & 36.56 & 53.55 & 30.59 & 46.85 & 69.95 & 82.32 & 41.45 & 55.69
  \\
  Zhang~\textit{et al.}~\cite{DA-CSLG} & 13.27 & 23.43 & 57.65 & 73.14 & 56.99 & 72.27 & 35.87 & 52.80 & 29.77 & 45.88 & 65.44 & 79.11 & 43.17 & 57.77 
  \\
  Wang~\textit{et al.}~\cite{DA-FGUDA} & 10.84 & 17.49 & 66.11 & 79.75 & 65.45 & 80.17 & 28.64 & 43.51 & 35.47 & 51.85 & 68.63 & 81.32 & 45.86 & 59.74 
  \\
  JDAF~\cite{JDAF} & \textbf{18.09} & \textbf{30.93} & 60.05 & 75.04 & 58.64 & 73.93 & 38.74 & 55.84 & 27.79 & 43.49 & 71.42 & 83.33 & 45.79 & 60.38
  \\
  CIA-UDA~\cite{DA-CIAUDA} & 9.20 & 16.86 & 53.39 & 69.61 & 63.36 & 77.57 & 44.90 & 61.97 & 43.96 & 61.07 & 70.48 & 82.68 & 47.55 & 61.63
  \\ 
  DAFormer~\cite{DAFormer} & 1.07 & 1.88	& 65.12 & 78.16 & 70.40 & 84.28 &	\textbf{61.25} & \textbf{76.59} & 49.02 & 65.51 & 82.44 & 89.70	& 54.88 & 66.02
  \\
  ST-DASegNet~\cite{ST-DASegNet} & 3.70 & 7.38 & 69.83 & 83.12 & \textbf{75.99} & \textbf{87.89} & 57.41 & 73.47 & 50.76 & 67.64 & \textbf{83.46} & \textbf{90.67} & 56.86 & 68.37
  \\ \cmidrule(r){1-15}
  \rowcolor{gray!30!} SAM-JOANet (ours) & 4.94 & 7.70 & \textbf{73.31} & \textbf{84.60} & 73.36 & 85.27 & 55.38 & 72.14 & \textbf{59.88} & \textbf{74.09} & 78.73 & 88.10 & \textbf{57.60} & \textbf{68.65}
  \\ \cmidrule(r){1-15}
   \end{tabular}
   \end{adjustbox}
   \caption{UDA-RSSeg comparison results (\%) on `` Vaihingen IR-R-G $\rightarrow$ Potsdam R-G-B''.}
 \label{Tab4}
 \end{table*}
\begin{table}
\centering
\begin{adjustbox}{scale=0.85}
\begin{tabular}{ccccc}
 \cmidrule(r){1-5}
  Methods & Background & Road & Building & $mIoU$
  \\ 
  \cmidrule(r){1-5}
  AdaptSegNet~\cite{AdapSegNet} & 49.84 & 11.09 & 52.71 & 37.88
  \\
  AdvEnt~\cite{ADVENT} & 50.42 & 22.92 & 49.43 & 40.92
  \\
  CLAN~\cite{CLAN} & 53.85 & 23.25 & 44.37 & 40.49
  \\
  MaxSquare~\cite{MaxSquare} & 53.47 & 20.75 & 43.06 & 39.09
  \\
  BDCA~\cite{BDCA} & 53.88 & 20.31 & 52.97 & 42.39 
  \\
  Wang~\textit{et al.}~\cite{Wangetal-PC} & 53.01 & 23.00 & 55.43 & 43.82
  \\
  ColorMapGAN~\cite{ColorMapGAN} & 54.58 & 15.15 & 49.07 & 39.61 
  \\
  SAC~\cite{SAC} & 55.69 & 20.86 & 55.09 & 43.88
  \\
  DPL~\cite{DPL} & 53.52 & 24.51 & 56.22 & 44.75 
  \\
  Chen~\textit{et al.}~\cite{DA-UDA1} & 54.20 & 24.66 & 56.32 & 45.06 
  \\
  ST-DASegNet~\cite{ST-DASegNet} & 51.84 & \textbf{45.27} & 59.26 & 52.12
  \\ 
  \cmidrule(r){1-5}
  \rowcolor{gray!30!} SAM-JOANet (ours) & \textbf{59.54} & \textbf{46.80} & \textbf{64.56} & \textbf{56.96}
  \\
  \cmidrule(r){1-5}
\end{tabular}
\end{adjustbox}
\caption{UDA-RSSeg comparison results (\%) on ``Paris $\rightarrow$ Chicago'' task.}
\label{Tab5}
\end{table}
\subsection{Experimental Results}
\subsubsection{Comparison Experiments on ISPRS dataset.} As shown in Tab.~\ref{Tab1}$\sim$Tab.~\ref{Tab4}, we compare the SAM-JOANet's performance with previous methods on the aforementioned 4 UDA-RSSeg task.
\par For ``Potsdam IR-R-G $\rightarrow$ Vaihingen IR-R-G'' task, source and target domain images have obvious discrepancy on geographical landscapes. The comparison results are shown in Tab.~\ref{Tab1}. Compared to ST-DASegNet~\cite{ST-DASegNet}, SAM-JOANet respectively achieves 2.53\% and 2.31\% improvement on $mIoU$ and $mF$-score. Particularly on ``Clutter'' category, SAM-JOANet surpasses all previous methods.
\par For ``Vaihingen IR-R-G $\rightarrow$ Potsdam IR-R-G'' task, the domain-shift issue is similar to the first task. As shown in Tab.~\ref{Tab2}, SAM-JOANet surpasses previous SOTA method, ST-DASegNet~\cite{ST-DASegNet} by 1.56\% on $mIoU$ value and 1.72\% on $mF$-score. On ``Car'' category, SAM-JOANet shows obvious superiority over previous methods. 
\par ``Potsdam R-G-B $\rightarrow$ Vaihingen IR-R-G'' task suffers from sensor variations and landscape discrepancy, so this task is more complex than the above two tasks. Specifically, ``Tree'' is hard to distinguish, because the color of ``Tree'' is green in R-G-B images while red in IR-R-G images. In Tab.\ref{Tab3}, we find that SAM-JOANet achieves the best $IoU$ value and the second best $F1$-score on this category. For overall performance, SAM-JOANet also ranks top-1, which proves the generalized performance in more challenging situation.
\par ``Vaihingen IR-R-G $\rightarrow$ Potsdam R-G-B '' task faces the similar challenge as the third task. In Tab.~\ref{Tab4}, we find that SAM-JOANet gains 0.74\% and 0.28\% on $mIoU$ value and $mF$-score when compared to ST-DASegNet. Especially on ``Impervious surfaces'' and ``Low vegetation'' categories, SAM-JOANet surpasses previous methods by large margin.
\subsubsection{Comparison Experiments on CITY-OSM Dataset.} As shown in Tab.~\ref{Tab5}, we make comparison between our proposed SAM-JOANet with previous SOTA methods on ``Paris $\rightarrow$ Chicago'' task.  Obviously, SAM-JOANet outperforms all previous methods. Specifically, SAM-JOANet surpasses ST-DASegNet~\cite{ST-DASegNet} by 4.84\% in $mIoU$ value, thereby achieving new SOTA results. On three categories, SAM-JOANet all shows stronger performance.
\begin{table*}	
	\centering
    \begin{adjustbox}{scale=0.75}
    \begin{tabular}{ccccc|cccccccccccc|cc}
  \cmidrule(r){1-19} 
  \multicolumn{5}{c}{Methods} & \multicolumn{2}{c}{Clutter} & \multicolumn{2}{c}{Impervious surfaces} & \multicolumn{2}{c}{Car} & \multicolumn{2}{c}{Tree} & \multicolumn{2}{c}{Low vegetation} & \multicolumn{2}{c}{Building} & \multicolumn{2}{c}{Overall}
  \\ \cmidrule(r){1-19} 
  \rowcolor{gray!30!} \multicolumn{19}{c}{Potsdam IR-R-G $\rightarrow$ Vaihingen IR-R-G}
  \\ \cmidrule(r){1-19}
  Baseline & PS & F-Adv & L-Adv & ST & $IoU$ & $F1$ & $IoU$ & $F1$ & $IoU$ & $F1$ & $IoU$ & $F1$ & $IoU$ & $F1$ & $IoU$ & $F1$ & $mIoU$ & $mF$ 
  \\ \cmidrule(r){1-19}
  \checkmark & - & - & - & - & 10.45 & 24.87 & 59.88 & 74.69 & 18.14 & 33.48 & 62.30 & 76.82 & 39.04 & 57.56 & 73.25 & 84.91 & 43.84 & 58.72
  \\ \cmidrule(r){1-19}
  \checkmark & \checkmark & - & - & - & 17.82 & 32.85 & 65.20 & 77.84 & 32.65 & 48.68 & 66.74 & 80.76 & 47.93 & 66.08 & 77.16 & 86.71 & 51.25 & 65.49
  \\ \cmidrule(r){1-19}
  \checkmark & \checkmark & \checkmark & - & - & 30.73 & 42.11 & 66.83 & 77.93 & 35.28 & 49.74 & 65.85 & 78.43 & 51.59 & 67.90 & 79.37 & 87.04 &  54.94 & 67.19
  \\ \cmidrule(r){1-19}
  \checkmark & \checkmark & - & \checkmark & - & 24.08 & 36.98 & 68.63 & 79.12 & 36.72 & 51.03 & 64.35 & 78.07 & 47.06 & 64.51 & 81.13 & 88.43 & 53.66 & 66.36 
  \\ \cmidrule(r){1-19}
  \checkmark & \checkmark & \checkmark & \checkmark & - & 40.11 & 54.29 & 71.49 & 83.54 & 45.88 & 64.57 & 65.72 & 78.95 & 52.40 & 68.56 & 83.22 & 90.83 & 59.80 & 73.46
  \\ \cmidrule(r){1-19}
  \checkmark & \checkmark & \checkmark & \checkmark & \checkmark & 68.88 & 81.57 & 72.16 & 83.83 & 52.68 & 69.00 & 69.23 & 81.82 & 54.39 & 70.46 & 83.81 & 91.19 & \textbf{66.86} & \textbf{79.65}
  \\ \cmidrule(r){1-19}
  \rowcolor{gray!30!} \multicolumn{19}{c}{Vaihingen IR-R-G $\rightarrow$ Potsdam IR-R-G}
   \\ \cmidrule(r){1-19}
  \checkmark & - & - & - & - & 0.42 & 0.68 & 60.81 & 75.26 & 50.74 & 67.83 & 26.65 & 48.19 & 42.69 & 60.18 & 68.06 & 82.74 & 41.56 & 55.81
  \\ \cmidrule(r){1-19}
  \checkmark & \checkmark & - & - & - & 6.35 & 15.38 & 66.89 & 80.15 & 53.78 & 71.42 & 38.09 & 55.36 & 50.40 & 67.97 & 72.40 & 83.29 & 47.99 & 62.26
  \\ \cmidrule(r){1-19}
  \checkmark & \checkmark & \checkmark & - & - & 4.43 & 12.17 & 70.93 & 81.48 & 65.10 & 79.94 & 38.97 & 57.25 & 54.07 & 70.68 & 78.16 & 86.71 & 51.94 & 64.70
  \\ \cmidrule(r){1-19}
  \checkmark & \checkmark & - & \checkmark & - & 2.70 & 7.91 & 69.45 & 81.34 & 67.44 & 82.58 & 36.97 & 54.87 & 53.04 & 70.24 & 73.29 & 85.59 & 50.48 & 63.76
  \\ \cmidrule(r){1-19}
  \checkmark & \checkmark & \checkmark & \checkmark & - & 5.12 & 8.69 & 74.25 & 84.01 & 72.99 & 83.27 & 49.37 & 65.62 & 55.18 & 71.85 & 82.80 & 89.63 & 56.62 & 67.18
  \\ \cmidrule(r){1-19}
  \checkmark & \checkmark & \checkmark & \checkmark & \checkmark & 5.72 & 8.44 & 75.97 & 85.78 & 74.24 & 85.35 & 68.62 & 80.91 & 58.35 & 75.61 & 84.37 & 90.18 & \textbf{61.21} & \textbf{71.05}
  \\ \cmidrule(r){1-19}
   \end{tabular}
   \end{adjustbox}
   \caption{Ablation study with category-level segmentation results on ``Potsdam IR-R-G $\rightarrow$ Vaihingen IR-R-G'' and ``Vaihingen IR-R-G $\rightarrow$ Potsdam IR-R-G'' (\%). ``PS'', ``ST'' respectively indicates prompted segmentor and self-training. ``F-Adv'' and ``L-Adv'' respectively indicate feature-level and logits-level adversarial learning. $F1$ and $mF$ respectively denote $F1$-score and $mF$-score.}
 \label{Tab6}
 \end{table*}
\subsubsection{Ablation Study.} To separately show the performance of each key component of SAM-JOANet, we conduct ablation study on ``Potsdam IR-R-G $\rightarrow$ Vaihingen IR-R-G'' and ``Potsdam R-G-B $\rightarrow$ Vaihingen IR-R-G'' adaptation tasks. As shown in Tab.~\ref{Tab6}, we will analyze in the following aspects. 
\par (1) \textbf{Baseline vs SAM-JOANet.} In this paper, the baseline model is designed as ``SAM-based Finetuning'' paradigm (Fig.~\ref{Fig1}(b)). Compared to baseline models on two tasks, the SAM-JOANet respectively improves by 23.02\% and 19.65\% on $mIoU$ value and 20.93\% and 15.24\% on $mF$-score. (2) \textbf{Effectiveness of prompted segmentor.} As shown in Tab.~\ref{Tab6}, the baseline model with a prompted segmentor shows significant improvement, indicating that the fine-tuning decoder benefits from the guidance of the prompted class-agnostic map. (3)  \textbf{Effectiveness of adversarial learning.} In SAM-JOANet, we integrate feature-level adversarial learning into the prompted segmentor. Comparative results demonstrate that feature-level adversarial learning enhances the segmentor. Logits-level adversarial learning, embedded in the fine-tuning decoder, also contributes to performance gains by reducing the discrepancy between features and predictive logits. (4) \textbf{Effectiveness of self-training.} Results in Table~\ref{Tab6} reveal that models employing self-training achieve further enhancements. This confirms that the self-training mechanism corrects the representation bias and stabilizes optimization during adversarial training, resulting in improved segmentation performance.
\par As shown in Tab.~\ref{Tab6}, we further provide the following specific analysis. (1) When integrating feature-level adversarial learning into the ``Baseline + Prompted Segmentor'', the model exhibits significant improvement. This enhancement occurs because feature-level adversarial learning boosts the predictive capabilities of the prompted segmentor. In turn, this enhanced prediction contributes to the refinement of the finetuning decoder by supplying a high-quality prompted map. (2) From Tab.~\ref{Tab1}, it is evident that feature-level and logits-level adversarial learning affect various categories differently. Feature-level adversarial learning is particularly beneficial for categories with fewer samples, enhancing their performance. Conversely, logits-level adversarial learning shows improved effectiveness in categories with a larger number of samples. Consequently, combining these two approaches results in a complementary and compatible two-scheme adversarial learning strategy that performs exceptionally well. (3) The self-training mechanism also demonstrates strong compatibility with adversarial learning, enhancing the model in a straightforward way. Essentially, it helps correct representation bias and stabilizes the optimization process during adversarial training. (4) In the ``Vaihingen IR-R-G $\rightarrow$ Potsdam IR-R-G'' task, we observe that our proposed key components consistently struggle with the ``Clutter'' category. This issue stems from the scarcity of training samples featuring ``Clutter''. This scarcity exemplifies how the few-shot problem adversely affects the performance in UDA-RSSeg task.
\begin{figure*}
\begin{center}
   \includegraphics[width=1.0\textwidth]{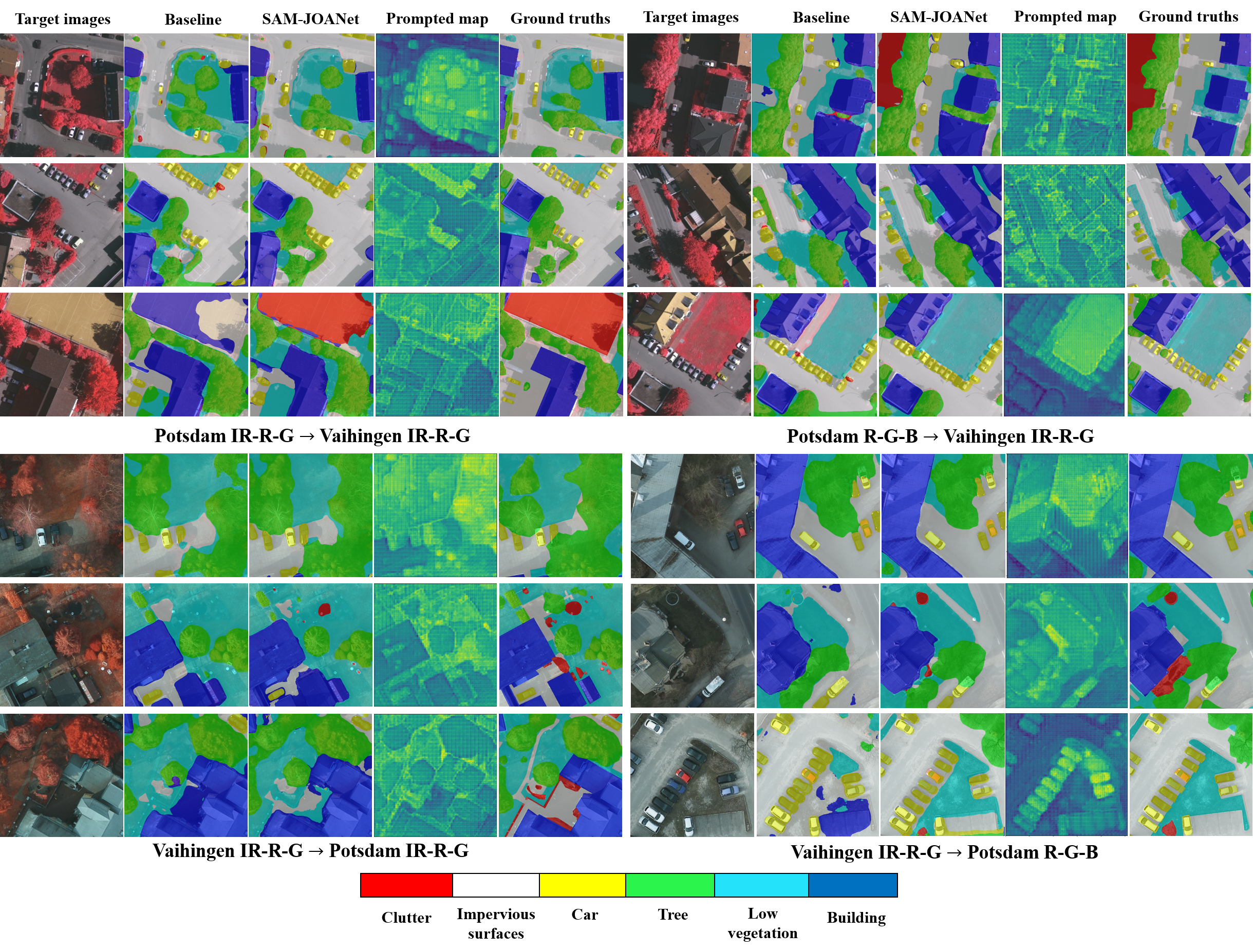}
\end{center}
   \caption{Visualization results of the 4 UDA-RSSeg task on ISPRS dataset.}
\label{Fig4}
\end{figure*}
\subsubsection{Visualization and Analysis}
To intuitively show the performance and interpretability of SAM-JOANet, we conduct visualization experiments on ISPRS and CITY-OSM datasets.
\par \textbf{Qualitative Visualization Results on ISPRS.} Besides the visualization and analysis shown in the main manuscript, we provide more qualitative visualization results on all 4 tasks (Fig.\ref{Fig4}). The abundant visualization on segmentation prediction intuitively prove the effectiveness of SAM-JOANet. With the guidance of high-quality class-agnostic map, SAM-JOANet shows the strong overall performance. In some specific category, SAM-JOANet can sometimes provide surprising results. Moreover, detailed visualization results coincides with the segmentation results shown in Tab.1$\sim$Tab.4 of main manuscript.
\par
\begin{figure}
\begin{center}
   \includegraphics[width=1.0\linewidth]{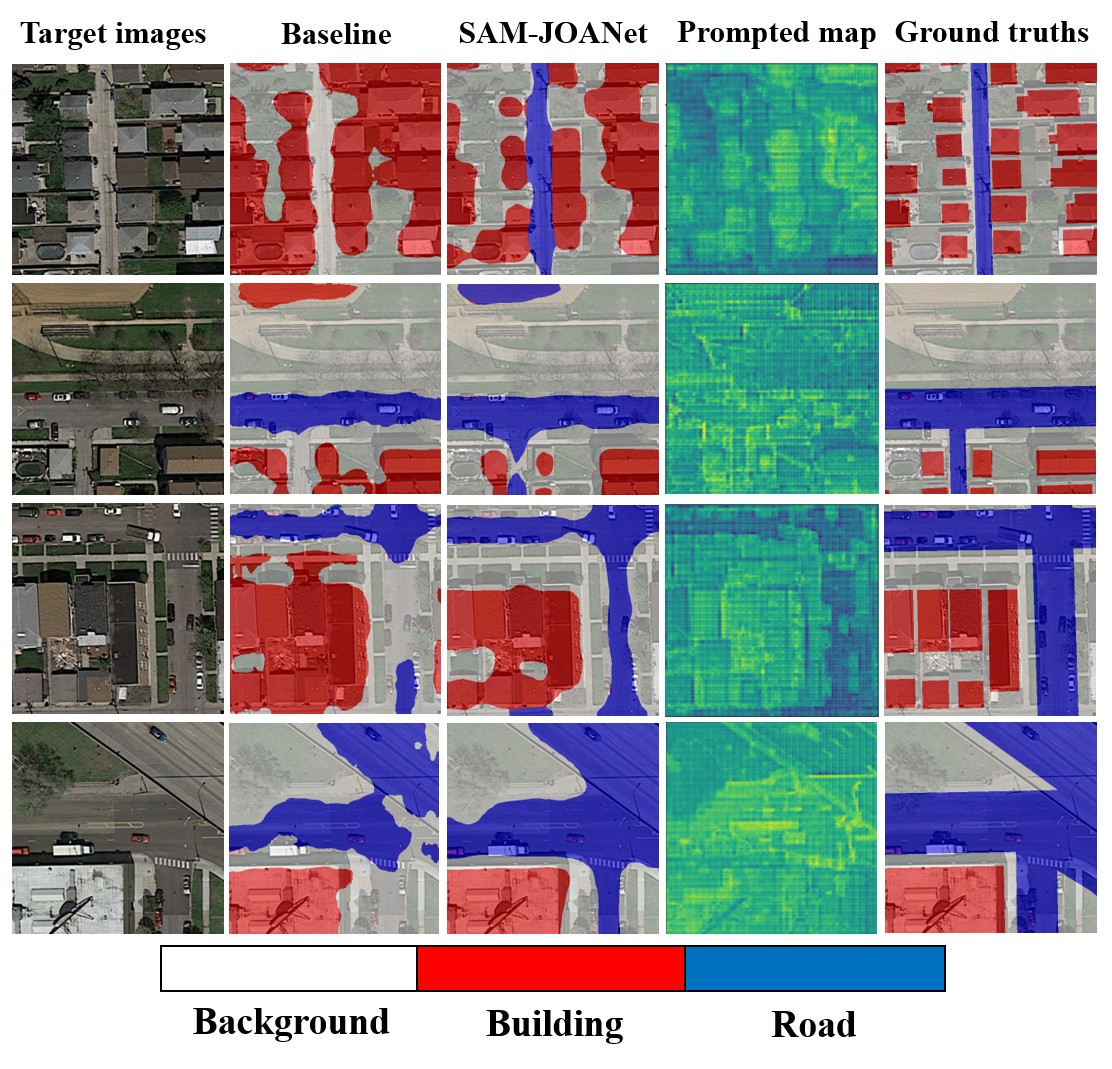}
\end{center}
   \caption{Visualization results of the ``Paris $\rightarrow$ Chicago'' task UDA-RSSeg task on CITY-OSM dataset.}
\label{Fig5}
\end{figure}

 \textbf{Qualitative Visualization Results on CITY-OSM.} To intuitively show the performance of SAM-JOANet on ``Paris $\rightarrow$ Chicago'' UDA-RSSeg task, we provide visualization results shown in Fig.\ref{Fig5}. From the results, we analyze in the following points: (1) Compared to baseline models, SAM-JOANet show obvious overall superiority, especially on ``Road'' category, which coincides with the results shown in Tab.5 of main manuscript. (2) It is clear that the prompted map shows clear boundary and accurate attention, which provides SAM-JOANet with important guidance. (3) From the results of baseline model (``SAM-based Finetuning'' paradigm), we find that even with generalized feature representation capability, it is still hard to overcome the domain gap when mapping feature to predictive logits. In this paper, we address this issue and propose SAM-JOANet to provide a insight for UDA-RSSeg task.
\begin{figure}
\begin{center}
   \includegraphics[width=1.05\linewidth]{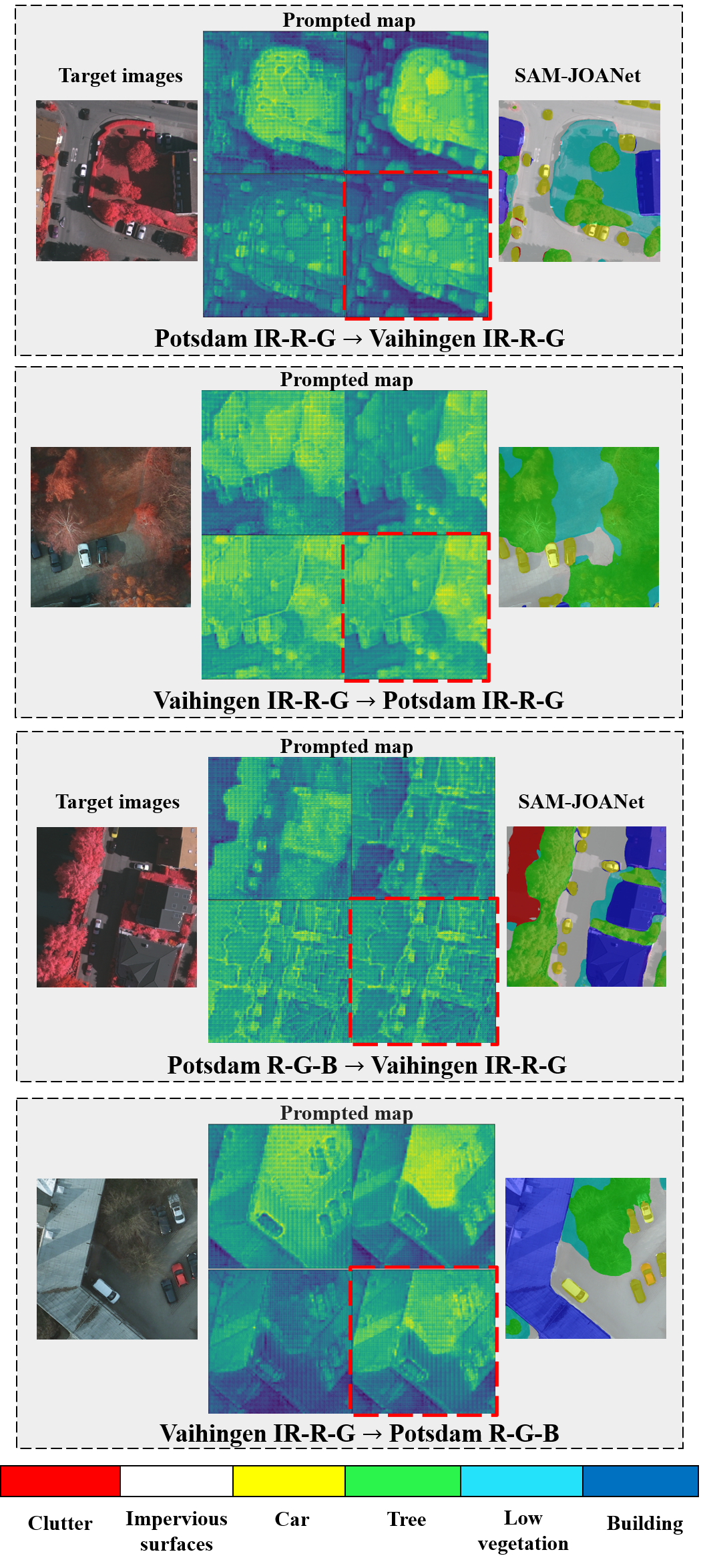}
\end{center}
   \caption{Visualization Analysis on Prompted Maps. The prompted maps without red doted boxes indicate the 3 prompted maps generated by SAM. The prompted maps with red doted boxes indicate the integrated prompted map with channel-wise $mean$ operation, which is used for guidance.}
\label{Fig6}
\end{figure}
\par \textbf{Visualization Analysis on Prompted Maps.} As show in Fig.\ref{Fig6}, we provide detailed visualization analysis on prompted maps. Here, we will analyze in the following aspects. (1) When guided by mask, SAM can provide 3 prompted maps with different guidance. In Fig.\ref{Fig6}, the images in ``left-top'', ``right-top'' and ``left-bottom'' are 3 prompted maps. (2) Obviously, 3 prompted maps have different attention tendency. Different prompted maps may focus on different categories or different instances. To maximally utilize the attention information of 3 prompted maps, we adopt channel-wise $mean$ operation to generate an integrated prompted map for guidance. The image in ``left-bottom'' indicates the integrated prompted map. (3) The segmentation results coincides with the attention tendency of prompted maps.
\section{Conclusion}
In this paper, we propose a joint-optimized adversarial network based on prompted foundation (SAM) to tackle UDA-RSSeg task. To essentially improve the feature consistency representation, we integrate SAM into our architecture to leverage its generalized representation capabilities. To bridge the domain gaps between features and predictive logits, we employ a feature-level adversarial prompted segmentor for SAM to generate prompted class-agnostic map as guidance. We also involve the logits-level adversarial learning and self-training mechanism for finetuning decoder to maximally boost the optimization effectiveness. Compared to previous paradigms, we first propose a ``SAM-based Joint-Optimization'' paradigm and provide an insight on UDA-RSSeg task. Extensive experiments on multiple benchmark datasets and visualization analysis show that SAM-JOANet achieves SOTA results in an interpretable manner.
\section*{CRediT Author Statement}
\textbf{Shuchang Lyu}: Conceptualization, Methodology, Software, Writing - Original Draft. \textbf{Qi Zhao}: Supervision, Project administration, Funding acquisition. \textbf{Guangliang Cheng}: Supervision, Methodology, Writing - Review \& Editing. \textbf{Yiwei He}: Validation, Visualization. \textbf{Zheng Zhou}: Investigation, Writing - Review \& Editing. \textbf{Guangbiao Wang}: Formal Analysis, Data Curation. \textbf{Zhenwei Shi}: Supervision, Writing - Review \& Editing.
\section*{Acknowledgment}
\label{sec:acknowledgment}
This work was supported by the National Natural Science Foundation of China (grant numbers 62072021).
\bibliographystyle{elsarticle-num} 
\bibliography{refs}

\end{document}